# Human Body Digital Twin: A Master Plan

Chenyu Tang[1,2], Wentian Yi[2], Edoardo Occhipinti[3], Yanning Dai[2], Shuo Gao[2*], and Luigi G. Occhipinti[1*]

[1]Department of Engineering, University of Cambridge
[2]School of Instrumentation Science and Optoelectronic Engineering, Beihang University
[3]UKRI Centre for Doctoral Training in AI for Healthcare, Department of Computing, Imperial College London
* Corresponding authors: lgo23@cam.ac.uk, shuo_gao@buaa.edu.cn

**Abstract:** A human body digital twin (DT) is a virtual representation of an individual's physiological state, created using real-time data from sensors and medical test devices, with the purpose of simulating, predicting, and optimizing health outcomes through advanced analytics and simulations. The human body DT has the potential to revolutionize healthcare and wellness, but its responsible and effective implementation requires consideration of multiple intertwined engineering aspects. This perspective article presents a comprehensive overview of the current status and future prospects of the human body DT and proposes a five-level roadmap to guide its development, from the sensing components, in the form of wearable devices, to the data collection, analysis, and decision-making systems. The article also highlights the necessary support, security, cost, and ethical considerations that must be addressed in order to ensure responsible and effective implementation of the human body DT. The article provides a framework for guiding development and offers a unique perspective on the future of the human body DT, facilitating new interdisciplinary research and innovative solutions in this rapidly evolving field.

## I. Introduction

Modern healthcare makes use of cutting-edge biomedical and nanomedicine technologies to deliver early prevention, accurate diagnoses, and precise treatments. Examples include early warning for neurodegenerative disease [1, 2], imaging-based diagnosis of cardio-cerebrovascular diseases [3, 4], and tumor-targeting therapies [5, 6]. Further progress is hindered by human bodies' uncertainty, originating from the complex relationships among organs, the unclear effects of everyday environment on human bodies, and the heterogeneity of different individuals. For example, the mechanisms behind the development of diseases like Amyotrophic Lateral Sclerosis (ALS) [7] and infant asthma [8] remain unclear because it's difficult to determine the precise effects of various external environmental factors and intrinsic factors (such as genes and the microbiome) on human physiological conditions. In these non-limiting case studies, as in many others, novel investigational approaches are needed, to unleash the ability of decoding the specific mechanisms of action leading to a disease and overcoming human bodies' uncertainty. Among these approaches, the idea of establishing and analyzing virtual representations of human organs and functions is burgeoning, supported by the exponential growth of computational capacity.

digital twin (DT) technology offers an opportunity to determine and predict the status of complex dynamical systems, even when conditions change. It explains that DTs are virtual replicas or representations of physical objects and that this technology can potentially decode uncertainties within the target system by utilizing sensors to collect extensive information over an extended period and leveraging advanced AI algorithms.

In this context, digital twin (DT) technology offers an opportunity to determine and predict the status of complex dynamical systems, even in the presence of changing conditions. DTs are virtual replicas or representations of physical objects. This technology has the potential to decode uncertainties within the target system by utilizing sensors to collect extensive information over an extended period and leveraging advanced AI algorithms. DTs have already demonstrated successful applications in diverse complex industrial scenarios. For example, in the manufacturing sector, they enhance production lines, equipment performance, bottleneck identification, equipment failure prediction, and optimize various processes by utilizing real-time data monitoring and integration with relevant data sources. Similarly, in transportation, DTs optimize delivery schedules, route planning, and fuel efficiency by integrating vehicle data, traffic sensors, and weather forecasts. These advancements have also sparked interest in applying DTs to the human body [9-11]. Nevertheless, the lack of a unifying approach, especially a consensus on the taxonomy and future roadmap of the human body DT, limits the development and deployment of human body DTs for use in healthcare.

Based on the analysis of state-of-the-art and technological developments in the field, we anticipate that future human body DTs will be refined into five levels that will play a unifying role in different healthcare applications. Therefore, to facilitate scientific and technological advancement in this promising field of engineering, we present here a perspective five-level blueprint for modeling the human body DT (Figure 1). The 5-level roadmap will serve as a convenient and unifying framework to establish a common language and facilitate collaboration among researchers in different fields. As human digital twin (DT) technology advances towards the development of level 5, which involves explainable models, personalized medicine is expected to benefit significantly from these advancements. By leveraging accurate and individualized models of human physiology, human body DT holds immense potential for accurately and swiftly predicting treatment outcomes prior to real-world implementation. This capability effectively reduces the risk of adverse reactions and streamlines the regulatory pathway in clinical trials. Furthermore, we discuss the necessary support and remaining issues that need to be addressed for the deployment of human body DTs, including security, cost, and ethical considerations.

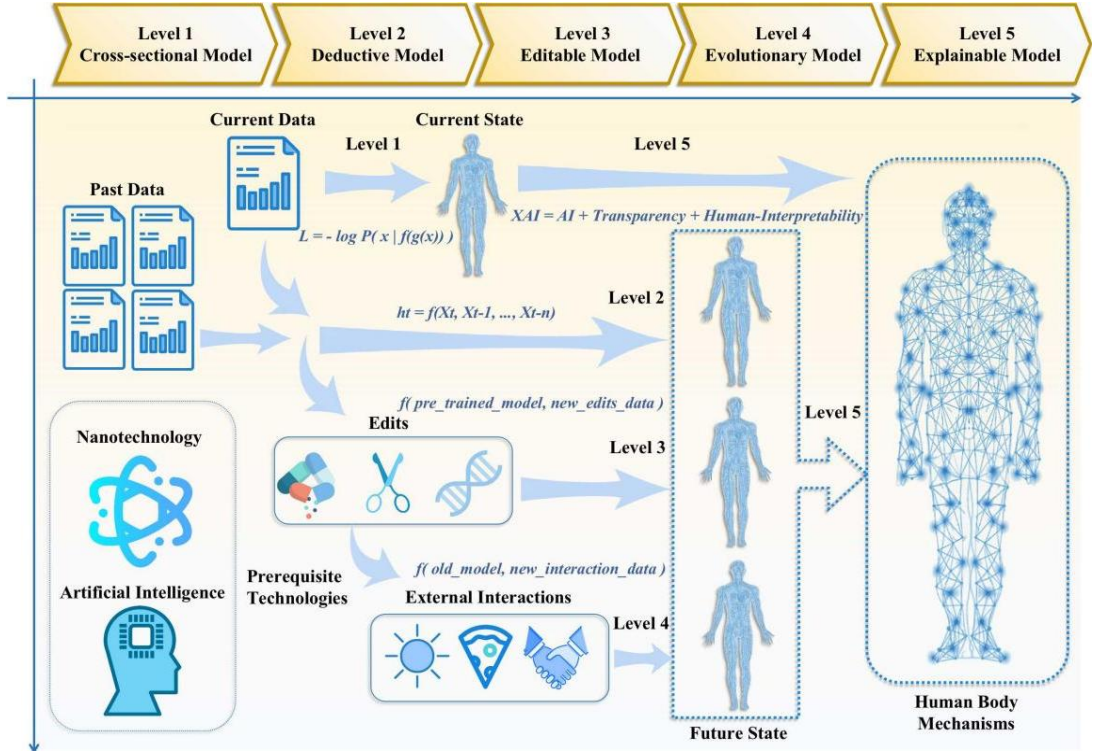

**Figure 1. The five-level roadmap for human body DT.** Level 1 (Cross-sectional Model) aims to determine various health indicators of the human body by using Artificial Intelligence (AI) classification methods on real-time data. Level 2 (Deductive Model) builds on the first level by incorporating not only real-time data but also past data to train predictive models for forecasting future health conditions. Level 3 (Editable Model) extends the previous levels by considering human interventions such as drug treatments, organ transplants, and gene editing to analyze and predict the health outcomes post-intervention. Level 4 (Evolutionary Model) adds another layer by also taking into account the interactions between the human body and the external environment, such as solar exposure, diet intake, and interpersonal connections, for more accurate health projections. Level 5 (Explainable Model) is based on the previous four levels and employs Explainable AI methods to delve into the underlying logic of health analysis and prediction. It explores not just the output results from data input to health state but also the biological principles involved.

## II. Five-level roadmap and related modelling methods

The scarcity of human data, especially data annotated by clinicians, poses a significant challenge to the advancement of the digital health field [10]. The development of a human body DT model is no exception to this issue. However, recent breakthroughs in related technologies offer promising solutions. Specifically, innovations in nanotechnology have facilitated the design and fabrication of novel sensors that are more sensitive, adaptable, and comfortable, thereby enabling large scale collection of human data over extended periods [12, 13]. Moreover, the advent of advanced Self-supervised Learning (SSL) algorithms allows for the utilization of copious amounts of unlabeled human data, a previously inconceivable accomplishment (see Box 1). SSL algorithms, combined with large-scale pre-training methods, have proven effective in fields like computer vision and natural language processing. AI models such as DALL·E 2 (which can complete intricate drawings based solely on descriptive sentences) [14] and ChatGPT (which can answer various complex questions, including coding and finance strategies) [15] pre-train on immense amounts of unlabeled data and fine-tune on small, labeled datasets for diverse tasks. The capacity to leverage unlabeled data in this manner aligns well with the demands of human body healthcare and holds immense potential for addressing the issue of insufficient human data for the development of a robust human body DT model. Under the impetus of these technologies (Figure 2), human body DT technology is anticipated to progressively unveil the mysteries of the human body along the five-level modeling roadmap outlined below.

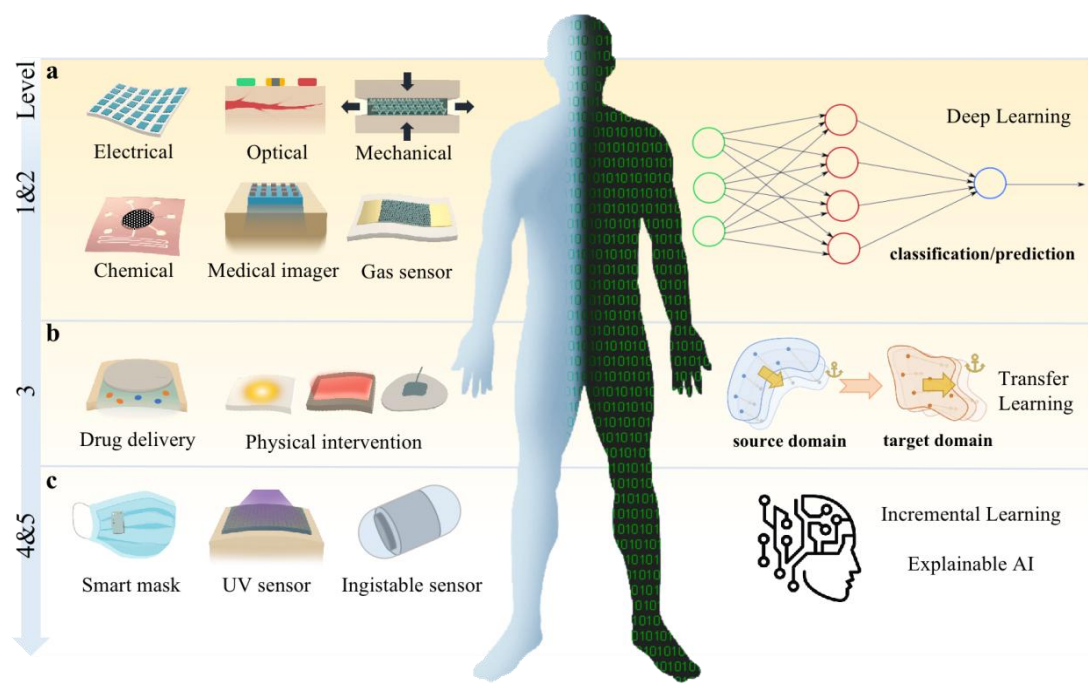

**Figure 2. Must have technologies to build human body DTs: left) the sensing devices to capture the human body data; right) the algorithms to model the human body.** The development stages of human digital twin pose different requirements for wearable devices. a) Level 1&2: Multimodality is crucial to measure signals of different nature and combine their information to extract complex patterns. Types of wearable sensors as the foundation of human digital twin, including traditional electrical, optical, mechanical, chemical sensors, as well as emerging sensors such as medical imaging devices, gas sensors, etc. b) Level 3: Actuators that can provide quantifiable outputs, such as targeted drug delivery devices, physical intervention devices like heat, light, electrical stimulations, etc. c) Level 4&5: Devices that monitor long-term exposure effects, such as smart masks that monitor air pollution, textile-based UV sensors, ingestible sensors that monitor digestive tract exposures, etc. On the right side the algorithms corresponding to each stage.

**Level 1 - Cross-sectional Model:**

In level 1, cross-sectional models are created for depicting the human body's digital portrait, by collecting data from the human body in a temporal cross-section to determine the real-time physical and biochemical states. Examples are locomotion classification [16] and metabolism monitoring [17]. The work in level 1 is an essential building block for the following levels.

Contrastive learning, as one of the most efficient algorithms among SSL algorithms in recent years, utilizes the intrinsic relationships between data modalities as pseudo-labels to train models (for example, DALL·E 2 uses the pair relationships between the images and their captions) [18]. This idea can be applied to the human body, where various multi-modal sensor data of the human body also have diverse intrinsic relationships. For instance, when monitoring human motion, sensors deployed at different locations produce various datasets, all corresponding to the same action or posture; when detecting neurodegenerative diseases, sensors such as inertial motion units, electroencephalography (EEG), electromyography (EMG) electrodes, and biochemical sensors for detecting disease-related biomarkers in biological fluids, such as saliva or sweat, produce patient-specific data [19]. Although these outputs have distinct patterns, they often contribute to map the same disease. These intrinsic relationships derived from the human body can be utilized as pseudo-labels to perform large-scale pre-training and facilitate the training of the Foundation Model (FM), see Box 1. These FMs can extract cross-sectional conditions of the human body, primarily through encoders that extract human information from different data modalities and can even zero-shot (without any further supervised training) decipher the cross-sectional status of the human body. More significantly, they will serve as the cornerstone of human DT models and continue to play a role in subsequent higher-level tasks.

**Level 2 - Deductive Model:**
Level 2 models perform deductive reasoning on the future development of the human body based on the time-continuous cross-sectional model information. By integrating the data from past cross-sections and the current cross-section, deductive models can predict evolutionary trends in human health status and potential disease risks. Several level 2 models have been reported [20, 21]. Models at this level have an intrinsic limitation: the deductions are based only on existing cross-sectional data, while the uncertain interventions on the human body and changes in the external world make it hard to yield an accurate long-term prediction. Hence, we need to develop higher-level models.

In establishing the Level 2 model, numerous past cross-sectional data can be encoded by pre-trained FMs through zero-shot or few-shot fine-tuning and fed into models for analyzing temporal states, including models based on network backbones such as Recurrent Neural Networks (RNN), Long Short-Term Memory (LSTM), and Transformer networks [22]. This enables the development of predicting models of future states through inference.

**Level 3 - Editable Model:**
In level 3, editable models, which can predict the impacts of edits (such as drug intervention, organ transplantation, gene editing, etc.) on human bodies, are created and utilized. Currently, many researchers are experimenting the edits on the human body to develop ways of curing or preventing diseases; however, unclear side effects cause the edits to be effective only in the short term [23, 24]. As an example, in [23], the researchers successfully transplanted kidneys from genetically modified pigs into two brain-dead human recipients. However, these kidneys remained viable and functioning for only 54 hours, primarily due to hyperacute rejection. This highlights the pressing need for the development of an Editable Model, capable of analyzing and predicting potential unintended consequences after edits like the aforementioned one due to unclear side effects. This level aims to leverage the pre-training data on cross-sections and perform few-shot inference based on limited edits (input data) to help cure some rare diseases and fatal illnesses.

For the Level 3 Model, the fundamental aspect is the integration of actual human edit data as new channels with the prior levels of the model. In comparison to cross-sectional data of the human body collected through routine health monitoring, the availability of body data post-edits (such as drug intake, surgeries, gene therapies, etc.) is limited. Utilizing pre-trained encoders from the previous levels and incorporating them with a small number of edit inputs is a promising approach to creating an editable model. To incorporate the edit data into the model, the parameters of the cross-sectional data encoders can be fixed, leaving only the parameters of the encoder for the edits and the decoder to be trained. This reduces the number of parameters to be trained, effectively addressing the challenge posed by the limited quantity of edit data.

**Level 4 - Evolutionary Model:**
Models in previous levels focus on interpretations and predictions of human bodies themselves with few considerations of influences of external factors playing significant roles in most if not all cases. However, the human body is by no means an isolated system. Interactions with the outside world, including solar exposure, diet intake, and interpersonal connections, can have subtle but determinant impacts on the human body [25-27]. In level 4, models can merge external factors into previous tasks to evolve and enhance its prediction accuracy, therefore named evolutionary. Quantifying these external factors and incrementally feeding them to the learning machine to update the DT model is the focus of level 4.

For the Level 4 Model, interactions between the human and the external environment are incorporated into the model. Unlike previous cross-sectional sensor data and edits, these interactions are no longer limited to a one-time input to the model. A single moment of interaction may have a minimal impact on human health, but when these interactions are accumulated over time, they can have a considerable influence on the individual's overall health. Therefore, the core task of this level is quantifying this accumulation and feeding it into the model as additional parameters. Some factors, such as respiration and light exposure, can be quantified well by ambient sensors, while more complex interactions, such as human social interactions, require the assistance of embedded sub-models to convert them into digitized inputs for the model.

Although systematic research on Evolutionary Models is not yet fully established, recent studies have confirmed the feasibility of real-time monitoring of important channels of interaction between the human body and external environment, such as solar exposure and breath [28, 29]. These works laid a solid foundation for the establishment of Level 4 human body DTs.

**Level 5 - Explainable Model:**
In the first 4 levels, models may offer accurate predictions and estimations, but they operate in a black-box or gray-box manner, meaning their internal working mechanisms are either entirely opaque or only partially understood. This lack of transparency makes it challenging to discern the relationships between inputs and outputs, leaving researchers ill-equipped to navigate the inherent uncertainties associated with human physiology. Therefore, in level 5, models will inform the researchers of the logical connections between observed phenomena and their outcomes. State-of-the-art research work just started to deploy explainable models relevant to this task. For example, researchers developed translatable systems based on medical imaging to explain the information contained in Computer Tomography (CT) scans or Magnetic Resonance Images (MRI) [30]. Current works in level 5 are still in their infancy and unlikely to provide real guidance to clinicians [31], while with the continuous development of human body DTs, models in level 5 may be integrated with suitable datasets to mine deep into actual features of the human body, as witnessed in the AI domain [32, 33], pushing the boundaries of future healthcare interventions.

The development of a Level 5 Model requires a deep understanding and explanation of the underlying mechanisms behind previous levels of the human body DT model. To achieve this, the use of advanced model interpretability techniques, such as saliency maps, activation maps, and model distillation, can provide valuable insights into the model's decision-making process [34]. The application of causal inference algorithms can further illuminate the relationships between inputs and outputs, allowing for a better understanding of the model's internal workings [35]. Additionally, incorporating model-agnostic interpretability methods, such as Local Interpretable Model-agnostic Explanations (LIME), Grad-CAM and SHAP, can offer a comprehensive view of the model's behavior, enabling a thorough understanding of not only how, but also why the model arrived at its predictions [34]. By leveraging these cutting-edge techniques, the Level 5 Model serves as a tool for advancing our knowledge of the human body and enhancing the reliability and trustworthiness of human body DT models.

## III. Must have technologies to capture the human body data

Human body DT models rely on input data obtained from wearable sensors positioned in various body locations. For cross-sectional and deductive models in levels 1 and 2, we have identified four main categories of such sensors: *electrical* such as electrocardiography (ECG), EEG, EMG; *optical* such as photoplethysmography (PPG); *mechanical* such as MEMS microphones or accelerometers; and *chemical* such as biosensors to detect specific target biomarkers from biological fluids (e.g., sweat), allowing direct monitoring of physiological states to infer health trends and disease risks (Figure 2a). Several companies have already incorporated one or more of these sensors into commercial wearable products like smartwatches, but further effort is required to encourage widespread adoption and continuous health monitoring. These sensors should integrate seamlessly in our daily lives by being comfortable, miniaturized, discreet, and equipped with long-lasting energy sources. Besides, various on-body and in-body sensors are required to provide holistic information for higher level of human DT models. Level 3 editable models additionally require actual human edit data (Figure 2b). Carefully quantified edits by wearable actuators generate the data needed to train models to avoid side effects during long-term therapies. Furthermore, sensors for capturing environmental stimulus are required to be incorporated in level 4 and 5, allowing spontaneous monitoring of human and surrounding environment (Figure 2c). The form of implementation of these sensors needs to be designed carefully to support long-term imperceptible monitoring and user convenience, as in the case of textile-based sensors integrated into clothing.

Current on-body wearable sensors with form factors such as textile, patches and tattoos can access to diverse analytes with no or minimal invasiveness. For large-area monitoring, fiber and textile-based sensors can be seamlessly integrated into clothing to facilitate spatially distributed analysis of signals including strain, UV exposure, pH, metabolites, environmental pollutants or biomarkers in sweat across epithelial areas [36]. Meanwhile, skin-conformable patches and tattoo-like devices using biocompatible adhesives are ideal for monitoring of physical parameters such as temperature, pressure, strain, and biomarkers in sweat or interstitial fluids [37, 38]. Emerging bioadhesive ultrasound (BAUS) devices can provide ultrasound imaging of organs and anomalies beneath the skin, enabling early diagnosis through viewing cardiac motility or vascular remodeling [39]. On the other hand, implantable sensors can provide direct transduction of biological signals under skin or in situ organs. For instance, continuous glucose monitors (CGM) from Abbott, Dexcom and Medtronic are implanted subcutaneously to measure glucose levels in blood and interstitial fluid, providing large dynamic datasets inaccessible from conventional finger-prick tests [40].

Considering that people are largely used to wearing earbuds, the ears provide a very encouraging alternative location for 24/7 continuous monitoring. Unlike the arms, they maintain a constant proximity to vital signal sources like the brain, lungs, and heart. They also exhibit a high level of vascularity, greater modulation with breathing, and a faster response to $SpO_2$ drops [42]. As such, recently developed multimodal in-ear sensors [43] have demonstrated their capability to measure EEG [44], ECG [45], PPG [32], microphone and accelerometer signals [41], thus showcasing their potential for various application of human-body DT models.

One of the biggest challenges to fully exploit the potential of wearables for continuous monitoring lies in the presence of artefacts. Indeed, these devices are prone to motion artefacts, which frequently result in the discarding of entire epochs of recordings. Such artefacts should be identified, classified, and removed in real-time with low-latency hardware-integrated algorithms. To remove these artefacts, one potential approach is to use the signals from mechanical sensors to capture and model the artefact. These sensors will only record a signal which is correlated with the artefact but not with the physiological signal of interest. Then, the signals from the mechanical sensors can be used as reference signals in an adaptive filter with an adaptive noise cancellation (ANC) configuration. Some preliminary results [41] have shown that accelerometers can be used to remove low-frequency artefacts generated by full-body movements (e.g. walking), while microphones can be employed to remove higher frequency artefacts generated by the relative movement between the sensor and the skin.

The influx of multidimensional sensor data generated by human body DTs poses significant computational challenges. Cloud-based centralized computing built on conventional von Neumann architectures is energy-intensive and constrained by limited data transmission bandwidth. Neuromorphic computing offers a promising alternative pathway to efficiently process the massive datasets involved in modeling complex physiological systems [46]. Neuromorphic devices emulate the signal processing and learning capabilities of biological nervous systems through networks of artificial neurons and synapses. By processing data locally where it is generated and harnessing the massive parallelism and adaptability of brain-inspired architectures, neuromorphic systems can potentially analyze streaming multimodal sensor data in real-time at far lower energy costs compared to traditional computing [47]. Integrating memristive synapses into wearable systems creates an artificial nervous network directly in the wearable platform [48-50]. Although limitations remain in switching speeds and reliability, progress in memristive materials and fabrication approaches may soon overcome these hurdles [51]. On-node and edge computing configurations place neuromorphic processors directly adjacent to sensors, avoiding data transmission lags. This tight sensing-computing integration empowers rapid reflexive actions and real-time adaptive decision making on-body.

## IV. Prospective applications

The evolution of the human body DT technology has ushered in a new era in the realm of medicine. By generating virtual counterparts that replicate human anatomy and physiology, it offers a remarkable opportunity to comprehend and anticipate diverse physiological and pathological states in a highly individualized manner.

One of the most exciting applications of the human body DT is in the field of personalized medicine. As the technology advances to Level 5, i.e., an explainable model, it will enable the creation of accurate and individualized models of human physiology. This, in turn, will allow for the design of personalized therapies for various diseases, taking into account the unique characteristics and requirements of each patient. The proposed human body digital twin model uses wearable sensors, which are cost-effective and convenient compared to large medical instruments, as the hardware backbone, and uses large foundation models as the basis, which can greatly increase user acceptance and reduce the model development cost, thus increasing the possibility of taking human body digital twin technology out of the laboratory. It thus has the potential to be used in large-scale applications to accurately and rapidly predict or diagnose various chronic diseases based on a personalized profile of different users, and then to help simulate their response to diverse therapies during the treatment phase. Clinicians can predict outcomes based on models before they are implemented in the real world, which minimizes the risk of adverse reactions and increases the chance of success while shortening the regulatory pathway in clinical trials.

Employing human body DT in personalized medicine offers several advantages. Firstly, it ensures a superior degree of precision in treatment, as the virtual model can simulate a patient's response to diverse therapies and prognosticate the outcome before their adoption in clinics. This minimizes the risk of unfavorable effects and bolsters the probability of success. Secondly, personalized medicine that leverages human body DT technology has the potential to be cost-effective. By pinpointing the root cause of a disease, it can reduce dependance on trial-and-error approaches and the employment of costly and potentially detrimental drugs. Furthermore, by facilitating real-time monitoring of patients, it can help to detect potential complications early, allowing for timely and efficacious intervention.

In summary, the advancement of human body DT technology harbors the potential to revolutionize the sphere of medicine and deliver an unprecedented level of precision and personalization in treating various ailments. With continued progress and refinement, it is conceivable that human body DT will become an indispensable instrument in diagnosing and managing a wide array of medical conditions.

## V. Outlook

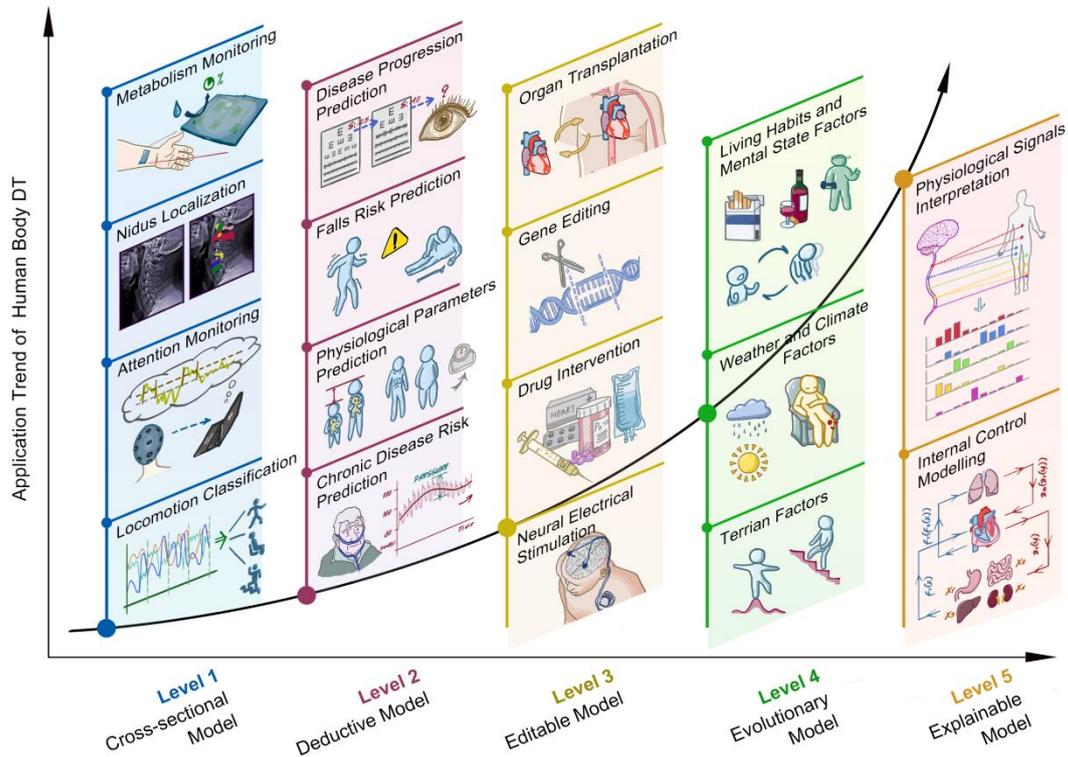

**Figure 3. Application trend of human body DT across various model levels.** DT models at Level 1 provide real-time assessments of human physical condition. As the model progresses to higher levels, it gains the ability to predict future physiological indicators (Level 2). At Level 3, the model undergoes editing and modification based on medical treatments. Additionally, at Level 4, the model incorporates the influence of external environmental factors. This progressive advancement culminates in Level 5, where the model achieves physiological signals interpretation and internal control modeling, enabling precise and personalized healthcare interventions.

This perspective provides a timely fresh look at the status and prospects of the rapidly evolving area of the human body DT, informing a master plan for the future development of the human body DT, stimulating discussion and new experimental approaches in this promising interdisciplinary domain.

Figure 3 provides a comprehensive depiction of the potential applications of the human body DT in healthcare and wellness, based on the proposed five-level roadmap. Following the proposed approach, human body DT models evolve from signal detection and real-time assessment (Level 1) to predicting physiological indicators and disease progression (Level 2), adapting to medical treatment (Level 3), and incorporating external factors (Level 4). Ultimately, at Level 5, the model aims to interpret body signals and internal functions, thereby facilitating deployment and adoption of personalized precision medicine. By progressively advancing along this roadmap, the envisioned human body DT model holds immense potential to revolutionize healthcare practices, enabling precise and tailored interventions based on comprehensive analyses performed in DT model before their use in human subjects.

We, therefore, anticipate a broader impact in future healthcare and in other research areas. For example, the adoption of human body DT in human assistive tasks for elders or physically impaired people may help alleviate the productivity deficit caused by global population aging and provide resilient solutions in the context of global economic slowdown cycles.

**Competing interests:** The authors declare no competing financial interests.

**Box 1: Self-Supervised Learning and Foundation Models in Healthcare Informatics**

**Unpacking Self-Supervised Learning (SSL)**

In machine learning paradigms, Self-Supervised Learning (SSL) is akin to decoding an intricate cipher without a predefined key. It ingeniously learns to make sense of data by leveraging inherent structures within the data itself, effectively "teaching" itself. This is a significant advantage in healthcare informatics, where annotated data are often scarce but the stakes are high. For instance, consider a repository of unlabelled Magnetic Resonance Imaging (MRI) scans. SSL algorithms can autonomously discern the anatomical and pathological features within these scans, thereby generating valuable knowledge without requiring explicit labels, a process that is both costly and time-consuming. In the context of Human Body DT, SSL uses data, not just from conventional sources like MRI scans, but also from around-the-clock wearable and implantable sensors. The method autonomously identifies vital signs and other physiological patterns. This approach is invaluable in healthcare informatics where constant monitoring is pivotal and large sets of annotated data are hard to come by.

**Foundation Models (FMs)**

Within the SSL framework, Foundation Models serve as the inaugural computational platform. These models are constructed by aggregating a broad range of unlabelled healthcare data, from traditional MRI scans to data harvested from modern wearable and implantable devices. This comprehensive approach creates a generalized foundation, facilitating the development of subsequent specialized predictive models tailored to various healthcare applications.

The significance of FMs lies in their versatility and adaptability. Initially designed to offer a broad understanding of healthcare data, they can be fine-tuned using minimal sets of labelled data. This allows them to be tailored for specific downstream applications like real-time health monitoring, diagnostic support, and early warning systems for emergent medical conditions.

**SSL in Healthcare: A Use-Case**

Expanding upon Foundation Models, wearables serve as a prime example of SSL's applicability in healthcare. These devices churn out a constant stream of unlabelled biometric data, such as heart rates and temperature readings. An FM, fine-tuned through SSL, can autonomously establish an individual's baseline physiological parameters. Once established, the model can alert healthcare providers to significant deviations that may warrant clinical investigation.

**Box 2: Implementation of Human Body DT in Healthcare**

The development of the human body DT has the potential to revolutionize healthcare. However, several considerations must be addressed to ensure that this technology is implemented responsibly and effectively.

**Necessary support**

Support from stakeholders is crucial. This includes data-sharing initiatives among healthcare organizations, securing government funding for research and development, nurturing interdisciplinary talent with expertise in fields such as biology, engineering, and computer science, and encouraging international collaborations to pool resources and expertise. For example, the DIGIPREDICT project (Grant No. 101017915) aims to seamlessly integrate various human body models and healthcare systems across Europe [52]. This project brings together experts from various fields to develop a collaborative platform for data sharing, and has been successfully deployed in several countries.

**Security**

Ensuring the security of patient data, from collection to storage, is essential due to the sensitivity of personal health information. This involves implementing robust data storage systems and encrypting data transmission, thus safeguarding against potential cyberattacks and data breaches. Restricting access to authorized personnel and granting patients control over their data usage are vital steps. Moreover, the connectivity of sensors exposes them to hacking, thereby posing risks on patients. Addressing these aspects is essential to maintain patient trust, as well as to comply with regulations such as the General Data Protection Regulation (GDPR) in Europe.

**Cost**

Implementing human body DT systems carries financial implications, including the cost of sensors, data processing and storage, energy consumption, and sustainability. Cost-effective and scalable solutions are pivotal to ensure broad accessibility across individuals and healthcare organizations. Ongoing research aims to explore low-cost scalable sensors and self-powered devices based on energy harvesting technology [53, 54]. These innovations have the potential to eliminate the need for frequent battery recharging or replacement, while extending the lifespan of these devices. Moreover, there are maintenance costs associated with model refinement to reflect the latest medical knowledge and technological advancements.

**Ethical issues**

Ethical considerations in human body DT implementation involve ensuring data anonymity, obtaining informed consent for data collection, and preventing misuse or discrimination. Equitable access is also essential to avoid exacerbating existing health disparities. Strategies such as government subsidies for low-income individuals, or the development of affordable, scalable solutions can make this technology widely accessible to everyone, regardless of their socioeconomic background.